\documentclass[conference]{IEEEtran}
\usepackage{times}

\usepackage{graphicx}
\usepackage{amsmath}
\usepackage{amssymb}
\usepackage[numbers]{natbib}
\usepackage{multicol}
\usepackage[bookmarks=true]{hyperref}
\usepackage{amsmath}
\usepackage[ruled,vlined]{algorithm2e}
\usepackage{subcaption}
\usepackage{balance}
\usepackage{comment}
\usepackage{xcolor}

\begin{document}

\title{Nightmare Dreamer: Dreaming About Unsafe \\ States And Planning Ahead}

\author{Author Names Omitted for Anonymous Review. Paper-ID [add your ID here]}

\author{%
\authorblockN{Oluwatosin Oseni$^{*1}$, Shengjie Wang$^{*2}$, Jun Zhu$^{2}$, and Micah Corah$^{1}$}
\authorblockA{
$^1$Colorado School of Mines
$^2$Tsinghua University\\
$^*$Equal contribution.
Corresponding author: \texttt{oluwatosin\_oseni@mines.edu}}
}


%

\maketitle

\begin{abstract}
Reinforcement Learning (RL) has shown remarkable success in real-world applications, particularly in robotics control. However, RL adoption remains limited due to insufficient safety guarantees. We introduce \textit{Nightmare Dreamer}, a model-based Safe RL algorithm that addresses safety concerns by leveraging a learned world model to predict potential safety violations and plan actions accordingly. \textit{Nightmare Dreamer} achieves nearly zero safety violations while maximizing rewards. \textit{Nightmare Dreamer} outperforms model-free baselines on Safety Gymnasium tasks using only image observations, achieving nearly a 20x improvement in efficiency.
\end{abstract}

\IEEEpeerreviewmaketitle

\section{Introduction}
\label{intro}

Reinforcement Learning (RL) has shown impressive success across various domains, from surpassing human performance in games like Go \cite{silver2017mastering}, to champion-level drone racing \cite{kaufmann2023champion}, and advanced robotic tasks like catching dynamic objects with dexterous hands \cite{lan2024dexcatchlearningcatcharbitrary}. Recent model-based RL methods, such as DreamerV3 \cite{dreamerv3}, further extend these capabilities—enabling robots to learn locomotion in hours and solving complex tasks like Minecraft diamond collection. Notably, RL has also demonstrated real-world impact in areas like plasma control for nuclear fusion \cite{tokmakarticle} and autonomous navigation of stratospheric balloons \cite{bellemare2020autonomous}.

Despite these successes, deployment of RL in real-world applications remains limited due to fundamental safety concerns. The exploratory nature of RL algorithms can lead agents to adopt dangerous or harmful behaviors during training, posing unacceptable risks in safety-critical environments \cite{feng2023dense}.

This challenge is notable when deploying RL agents in environments where they interact with or operate around humans, such as autonomous vehicles, robotic assistants, or industrial control systems.
Safe Reinforcement Learning (SafeRL) addresses these concerns by formulating the learning problem as a Constrained Markov Decision Process (CMDP) \cite{Altman1999ConstrainedMD}, where agents must maximize rewards while satisfying explicit safety constraints. Current approaches primarily rely on two methodologies: Lagrangian-based methods known as the primal-dual method \cite{minmax} that use dual optimization to balance rewards and constraints with algorithms like PPO-Lag and TRPO-Lag \cite{ray2019benchmarking}, and primal methods that attempt to apply the cost constraints with clever design of the objective functions as well as updating the policy without much use of dual variables \cite{xu2021crponewapproachsafe,cen2021fastglobalconvergencenatural}. State-of-art Model-free approaches like CPO \cite{achiam2017constrainedpolicyoptimization}
and
PPO-Lagrangian \cite{ray2019benchmarking}, while theoretically sound, suffer from sample inefficiency and struggle to maintain safety guarantees throughout training, particularly in high-dimensional visual environments. Conversely, model-based methods often fail to fully exploit the predictive capabilities of learned world models for proactive safety planning, limiting their effectiveness in preventing future constraint violations.
To address these limitations, we introduce \textit{Nightmare Dreamer}, a model-based SafeRL algorithm that leverages learned world models to predict potential safety violations and plan actions accordingly. Our key innovation lies in the integration of dual specialized actors—a control actor optimized for reward maximization and a safe actor focused on constraint satisfaction—with an online planning algorithm that switches between policies based on predicted future costs. 
Unlike existing approaches that treat safety as a reactive constraint, \textit{Nightmare Dreamer} proactively ``dreams'' about unsafe future states and takes preventive action.
Our main contributions are threefold: 
\begin{enumerate}
    \item  A bi-actor architecture that separates reward optimization from safety constraint satisfaction, enabling more effective multi-objective learning;
    \item  A predictive safety planning mechanism that uses world model rollouts to anticipate constraint violations; 
    \item  Demonstration that discriminator-based regularization can achieve stable training and superior performance compared to traditional behaviour cloning approaches.
\end{enumerate}
Experimental evaluation on Safety Gymnasium benchmarks demonstrates that \textit{Nightmare Dreamer} achieves nearly zero safety violations while maintaining competitive reward performance. Moreover, \textit{Nightmare Dreamer} demonstrates strong sample efficiency, surpassing baseline performance with as little as 1/20 of the interaction steps.

\section{Related Work}
\label{related work}
Safety will play a vital role in potential everyday adoption of RL. SafeRL seeks to address this challenge
by maximizing an objective function (reward) while simultaneously maintaining safety constraints (cost) below a predefined safety budget.

\textbf{Constrained Policy Optimization (CPO)}~\cite{achiam2017constrainedpolicyoptimization}, a primal method, was the first state-of-the-art policy gradient algorithm to solve the CMDP problem. CPO performs two policy updates: first, updating the policy in the direction of objective optimization (similar to Trust Region Policy Optimization (TRPO) \cite{schulman2017trustregionpolicyoptimization}), followed by projecting the policy back into the constraint set. While CPO may outperform primal-dual methods on some tasks and converge to the safety bound, it is computationally intensive due to being a second-order method involving inversion of high-dimensional Hessians \cite{zhang2022penalizedproximalpolicyoptimization}.

\textbf{Primal-dual methods} such as PPO-Lagrangian and TRPO-Lagrangian \cite{ray2019benchmarking} are the standard approach for CMDP problems. These methods apply Lagrangian duality in SafeRL and have achieved considerabe success. However, primal-dual methods remain challenging to apply due to parameter sensitivity in the learning rate of the Lagrangian multiplier.

\textbf{Model-based Vision-only Safe RL} Model-based methods have historically outperformed model-free approaches due to their superior sample efficiency. LAMBDA~\cite{as2022constrainedpolicyoptimizationbayesian} added safe planning capabilities to DreamerV1~\cite{hafner2020dreamcontrollearningbehaviors}, but suffers from the same suboptimal performance due to its base algorithm, DreamerV1, compared to its later improvement, DreamerV2. Safe SLAC~\cite{hogewind2022safereinforcementlearningpixels} achieves comparable performance to LAMBDA, yet does not fully exploit the world model's safety augmentation potential, bypassing imaginary rollouts that could enhance safe policy learning. Safe Dreamer~\cite{huang2024safedreamersafereinforcementlearning}, an adaptation of DreamerV3, combines Lagrangian methods with online planning to achieve strong performance. However, their planning is computationally intensive at inference time.
\section{Preliminary}
\label{Preliminary}

Safe RL tries to solve the Constrained RL problem known as Constrained Markov Decision Processes (CMDP)\cite{Altman1999ConstrainedMD}, an extension of MPD in classical RL. The CMDP can be represented in a tuple $\mathcal{M} =( \mathcal{S}, \mathcal{A}, \mathcal{R}, \mathcal{C}, \mathbb{P}, \mu, \gamma)$. $\mathcal{S}$ refers to the state space; $\mathcal{A}$ refers to the action space;  $ \mathcal{R}(r | s, a)$ refers to the reward obtained by the agent in state s after taking action $a$; $\mathcal{C}(c| s, a)$ refer to cost which will be subject to a constraint;  $\mathbb{P}(s'|s, a)$ is the transition probability of going to state $s'$ from state s taking action a while receiving $ \mathcal{R}(r | s, a)$ and $\mathcal{C}(c| s, a)$, $\mu: S \rightarrow [0,1]$ is the starting state distribution, and finally $\gamma \in [0,1)$ is the reward discount factor. We define a parametrized policy $\pi_\theta(a|s)$ to maximize the cumulative discounted reward defined in the objective function below:

\begin{align}
J^R(\pi_\theta) =  \mathbb{E} \biggr[ \sum_{t =0}^{\infty} \gamma \mathcal{R}(r | s, a)\biggl]
\end{align}
In SafeRL, we aim to learn a policy that maximises the above objective while satisfying the constraint:
\begin{align}
J^C(\pi_\theta) =  \mathbb{E} \biggr[ \sum_{t =0}^{\infty} \gamma \mathcal{C}(c| s, a)\biggl] \leq b.
\end{align}
In other words, we maintain cumulative discounted cost below a threshold known as the safety budget $b$.
We formulate the SafeRL problem as finding the optimal policy that satisfies:
\begin{align}
\pi_{*} = \underset{\pi_\theta}{\arg\max}~J^R(\pi_\theta)  \quad \text{s.t} \quad J^C(\pi_\theta)  \leq b.
\label{eq:constrained_optimization}
\end{align}

\section{Safe World Model Learning}

\textit{Nightmare Dreamer} learns a world model based on the work of
\citet{hafner2019learninglatentdynamicsplanning}, adopting the Recurrent State-Space Model. 
The world model components are parametrized by $\epsilon$ and learns the world dynamics. Following the partial-observability framework of RL problems, our model takes in the current camera image and the observation $o_t \sim p(o_t|h_t, z_t)$.
Then, without access to the real state, our model computes an internal state $h_t, z_t$ to learn the world dynamics as well as to predict future observations, rewards, and costs:
\begin{align*}
    \text{Recurrent Model: } & h_t = f_\epsilon \left(h_{t-1}, z_{t-1}, a_t \right) \\ 
    \text{Encoder Model: } & z_t \sim q_\epsilon \left(z_{t} \mid h_t, o_t \right) \\
    \text{Decoder Model: } & \hat{o_t} \sim p_\epsilon \left(o_{t} \mid h_t, z_t \right) \\
    \text{Transition Model: } & \hat{z_t} \sim p_\epsilon \left(\hat{z_t} \mid h_t \right) \\
    \text{Reward Model: } & \hat{r_t} \sim p_\epsilon \left(\hat{r_t} \mid h_t, z_t \right) \\
    \text{Cost Model: } & \hat{c_t} \sim p_\epsilon \left(\hat{r_t} \mid h_t, z_t \right) \\
    \text{Discount Model: } & \hat{\gamma_{t}} \sim p_\epsilon \left(\gamma_{t} \mid h_t, z_t \right).
\end{align*}
We define the loss function,
parametrized by $\epsilon$, below:
\begin{equation}
\begin{aligned}
    \mathcal{L}(\epsilon) \stackrel{.}{=}\sum\limits_{t=1}^{T}
    & -\alpha_c \underset{\text{cost log loss}}{ \ln(p_{\epsilon}(c_t | h_t, z_t))}  -\alpha_r \underset{\text{reward log loss}}{ \ln(p_\epsilon (r_t | h_t, z_t))} \\
  &  \underset{\text{reconstruction loss}}{ -\ln(p_{\epsilon}(o_t | h_t, z_t))} 
    \underset{\text{discount log loss}}{ -\ln(p_{\epsilon}(y_t | h_t, z_t))}. \\ 
    &  \underset{\text{representation loss}}{+\text{KL} \left[ q_{\epsilon}(z_t | h_t, o_t) \,||\,  sg(p_{\epsilon}(z_t | h_t) \right]}.
\end{aligned}
\end{equation}

\section{Bi-Actor Critic Learning}
We take a multi-agent Actor-Critic Model-based RL approach to tackle the SafeRL problem. We train a Control and Safe Actor and a Reward and Cost Critic in pairs and aim to optimize their respective policy and value functions separately. To optimize both Policy functions, we utilize the learned world model and perform trajectory rollouts under each policy. 
The rollouts involve sampling a real observation stored from a previous environment interaction and, with the learned dynamics, imagining future states by applying the sampled action from the actors for $H$ horizon steps.
However, during environmental interaction, we switch between the Control and Safe actor to ensure safety.

\subsection{Control Actor and Safe Actor}
In our framework, we define two actor types. The Control actor, parametrized by $\phi$, executes action $a_c$ sampled by $\pi_{\phi}(a|s)$ that maximizes future expected reward. On the other hand, the safe actor, parametrized by $\rho$, aims to find an action $a_s$ sampled by policy $\pi_{\rho}(a|s)$ that satisfies the cost constraint. 
Both actors are implemented as Multi-layer Perceptron (MLP) networks and employ the Exponential Linear Unit (ELU) activation function \citep{clevert2016fast}. Action predictions are modeled using a truncated normal distribution.

\subsection{Planning Ahead of Risks}
\begin{figure}
    \centering
    \includegraphics[width=\linewidth]{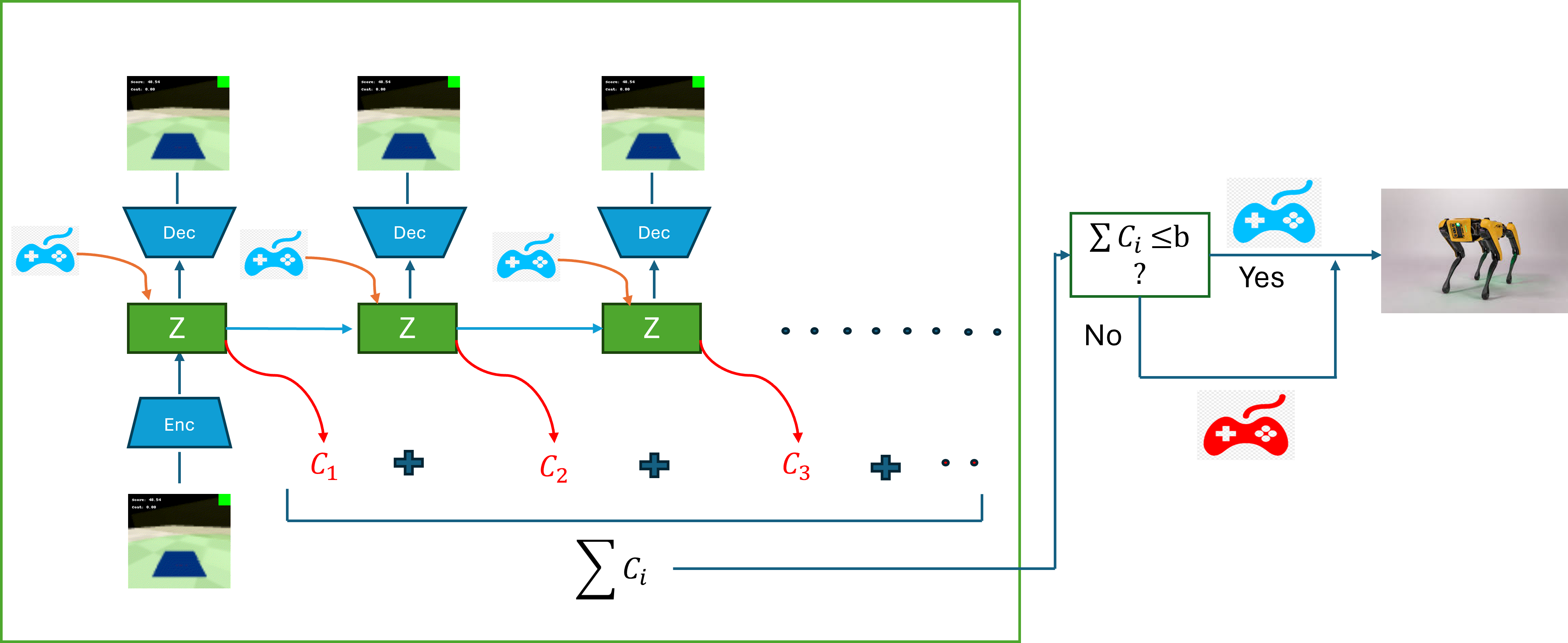}
    \caption{Action selection during environment interaction: The \textcolor{blue}{blue game pad} signifies the action from the Controller, while the \textcolor{red}{red game pad} refers to an action from from the Safe Actor.}
    \label{fig:action-selection}
\end{figure}

Obstacle avoidance is a fundamental concept in robotics and planning.
We pursue a similar approach (pseudo-code in Alg.~\ref{alg:risk_planning_simple}) by leveraging the learned world model. Starting from the current observation and state embedding, we perform rollouts under the Control Policy $\pi_{\phi}(a | s) $ in the latent space. This allows us to predict the potential cost violation if we continue to act under the Control policy. If the cost violation exceeds a threshold (safety budget $b$), we simply perform action selection by sampling from the safe policy $\pi_{\rho}(a|s)$ and sample from the Control policy $\pi_{\phi}(a|s) $ otherwise. 
We provide the action selection equation below:
\begin{align}
a &\sim (1 - \mathbf{1} (C_\mathrm{sum} \leq b_s)) \cdot \pi_{\rho}(a|s) + \mathbf{1} (C_\mathrm{sum} \leq b_s) \cdot  \pi_{\phi}(a|s) \nonumber \\
& \quad C_\mathrm{sum} = C_t(h_t,z_t,o_t) + \sum_{t+1}^{t+1+H} C_t(h_t, z_t) 
\end{align}
where $C$ at time step $t$ is predicted from the agent's observation (Posterior), and the consequent Costs are predicted using rollouts from the learned model (Prior) without access to the observation. Figure~\ref{fig:action-selection} illustrates this action selection process.

\subsection{Control \& Safe Actors Learning}

In our approach, we train two policies, a Control policy and a Safe policy. To train the Control policy, which actors aim to find a set of parameters that maximize the objective (Reward). We achieve this by directly maximizing the target Value function (gradient ascent) wrt the Control Policy, which is given by:
\begin{equation}
    V_t^{\lambda} = r_t + \gamma_t \begin{cases}
(1 - \lambda)v_\xi(\hat{z}_{t+1}) + \lambda V_{t+1}^{\lambda} & \text{if } t < H, \\
v_\xi(\hat{z}_H) & \text{if } t = H.
\end{cases}
\end{equation}
Where the value function $v_\xi$ aims to maximise the sum of discounted rewards.
Finally, we obtain the following loss function for the Control policy parametrized by $\phi$~\citep{dreamerv2}:

\begin{equation}
\mathcal{L}(\phi) \stackrel{.}{=} \sum_{t=1}^{H-1} (\underset{\text{target value}}{- V^{\lambda}_{t}} - \underset{\text{entropy regularizer}}{\eta H[\pi_{\rho}(a_t | s_t)]}).
\end{equation}

The entropy regulariser term ensures policy exploration. To perform policy improvement, we perform rollouts under the Control policy by sampling states from the buffer and performing H horizon steps from the sampled initial states. The rollouts (Imagination) involves starting from some initial state and, with the learned model (Transition Model), computing all recurring states given an Action.
\subsubsection{Safe Actor Learning and Lagrangian Formulation}
The Primal-dual method~\citep{minmax}, also known as the Lagrangian method, is the most common approach to the Multi-objective CMDP problem~\cite{ray2019benchmarking}. The Lagrangian formulation balances our task objective with cost constraints and is given below:

\begin{equation}
    \underset{\pi_\phi}{\max} \quad \underset{\lambda_{p} \geq 0}{\min} \quad J_\mathrm{task}(\pi_\phi) - \lambda_{p}(J_\mathrm{constraint}(\pi_\rho) - b).
\label{eq:primal_dual}
\end{equation}
The Safe actor, parametrized by $\rho$, aims to find a policy $\pi_\rho(a|s)$ that solves the multi-objective optimization problem \eqref{eq:primal_dual} and is given by:

\begin{equation}
\mathcal{L}(\rho) \stackrel{.}{=} \sum_{t=1}^{H-1} ( \underset{\text{target cost value}}{\lambda_{p} C ^ {\lambda}_{t}} \quad \underset{\text{behavior cloning}}{-D(a_t, s_t)}   - \underset{\text{entropy regularizer}}{\eta H[\pi_{\phi}(a_t | s_t)]}).
\label{eq:safe_policy_loss}
\end{equation}
The first term aims to solve the cost constraint in \eqref{eq:constrained_optimization} by directly minimizing the target value cost function wrt to the Safe Policy and is weighed by $\lambda_p$, provided in \eqref{eq:lagrangian_update}.

We achieve maximizing the control objective in \ref{eq:constrained_optimization} by \textbf{Behavior Cloning} the Control Policy $\pi_\phi(a|s)$ rather than direct reward maximization. This Multi-Objective loss function ensures the satisfaction of safety constraints while maintaining goal-directed behavior.

To imitate the Control policy, we train a Discriminator network that learns to predict if an action is sampled from a Control or a Safe policy, given a current state. The discriminator outputs 0 for safe actions and 1 for control actions, and we maximize $-D(s, a)$ wrt the Safe Policy to encourage control-like behavior.

\begin{algorithm}
\caption{Planning Ahead of Risks for Safe Action Selection}
\label{alg:risk_planning_simple}
\SetAlgoLined
\KwIn{Current state $s_t$, safety budget $b_s$}
\KwOut{Action $a_t$ to execute}

Compute current cost $C_t(h_t, z_t, o_t)$ based on current observation\;
Initialize $C_\mathrm{sum} \gets C_t(h_t, z_t, o_t)$\;

\For{$i \gets 1$ \KwTo $H$}{
    Predict next latent state using learned dynamics model\;
    Estimate cost $C_{t+i}$ for predicted state\;
    $C_\mathrm{sum} \gets C_\mathrm{sum} + C_{t+i}$\;
}

\eIf{$C_\mathrm{sum} > b_s$}{
    $a_t \sim \pi_{\rho}(a|s_t)$ \tcp*{Sample action from safe policy}
}{
    $a_t \sim \pi_{\phi}(a|s_t)$ \tcp*{Sample action from Control policy}
}

\Return{$a_t$}\;
\end{algorithm}

\subsubsection{Lagrangian Method and Update}

The learnable Lagrangian value $\lambda_{p}$ in Equ  \ref{eq:safe_policy_loss} is updated based with:
\begin{equation}
    \lambda_{p_{k+1}} = \text{Clip}( \lambda_{p_k} - \alpha ( C_k - \text{budget}), \lambda_{p_{\min}} , \lambda_{p_{\max}} )
\label{eq:lagrangian_update}
\end{equation}
where the Clip function ensures the Lagrangian value does not explode or become infinitesimal, and the online mean cost $C_k$ is defined as the moving average cost over the past $l = 50$ time steps.
This online mean provides an estimate of the current performance of the agent that allows us to update the Lagrangian multiplier $\lambda_p$ appropriately.

\subsubsection{Behavior Imitation via Action Discrimination}
Discriminators are often used with adversarial networks and GANs~\cite{gans} to perform image generation. We use a Discriminator term to ensure our safe policy selects actions that maximise the reward by mimicking the control policy.
Our experiments show that direct Discriminator optimization provides superior regularization and generalization compared to standard behavior cloning methods like KL loss or log probability. The discriminator estimates the log likelihood of predicting if an action is sampled from the Control policy or a safe policy given a current state $s_t$:

\begin{equation}
    \mathcal{L}(D(a_t, s_t))  \stackrel{.}{=}  \mathbb{E}_{a \sim \pi_{\phi}(a|s) }\log(D ( a_t, s_t)).
\end{equation}

\subsection{Critics Learning}
The reward and cost critics are trained to predict future discounted reward and cost, respectively, given the states from the imagination rollouts, similar to the actors. The Critics are MLP networks and use ELU activation functions that output a distribution of the critic estimation.
We found that this helps address the sparse nature of the cost from the environment.

Leveraging the learned model allows us to predict the sparse cost in the environment.
We compute the future discounted sum of both Cost and reward using the generalized $\lambda$ target values~\citep{schulman2018highdimensional,sutton2018reinforcement}. using the same setup for  DreamerV2 for both the Cost and Reward Value functions.

The loss functions are thus formulated to maximize the log-likelihood of predicting the value function from the $\lambda$ target values.

\section{Experimental Setup and Results}
\textit{Nightmare Dreamer} was trained on Safety Gymnasium, a Safe Policy Optimization Benchmark (SafePO) for Safe RL \citep{ji2023safety}. This benchmark is an extension of the now-discontinued Safety Gym previously maintained by OpenAI \citep{fujimoto2019benchmarkingbatchdeepreinforcement}.

\subsection{Task Description}
\begin{figure}
    \centering
    \begin{subfigure}{0.3\linewidth}
        \centering
        \includegraphics[width=\linewidth]{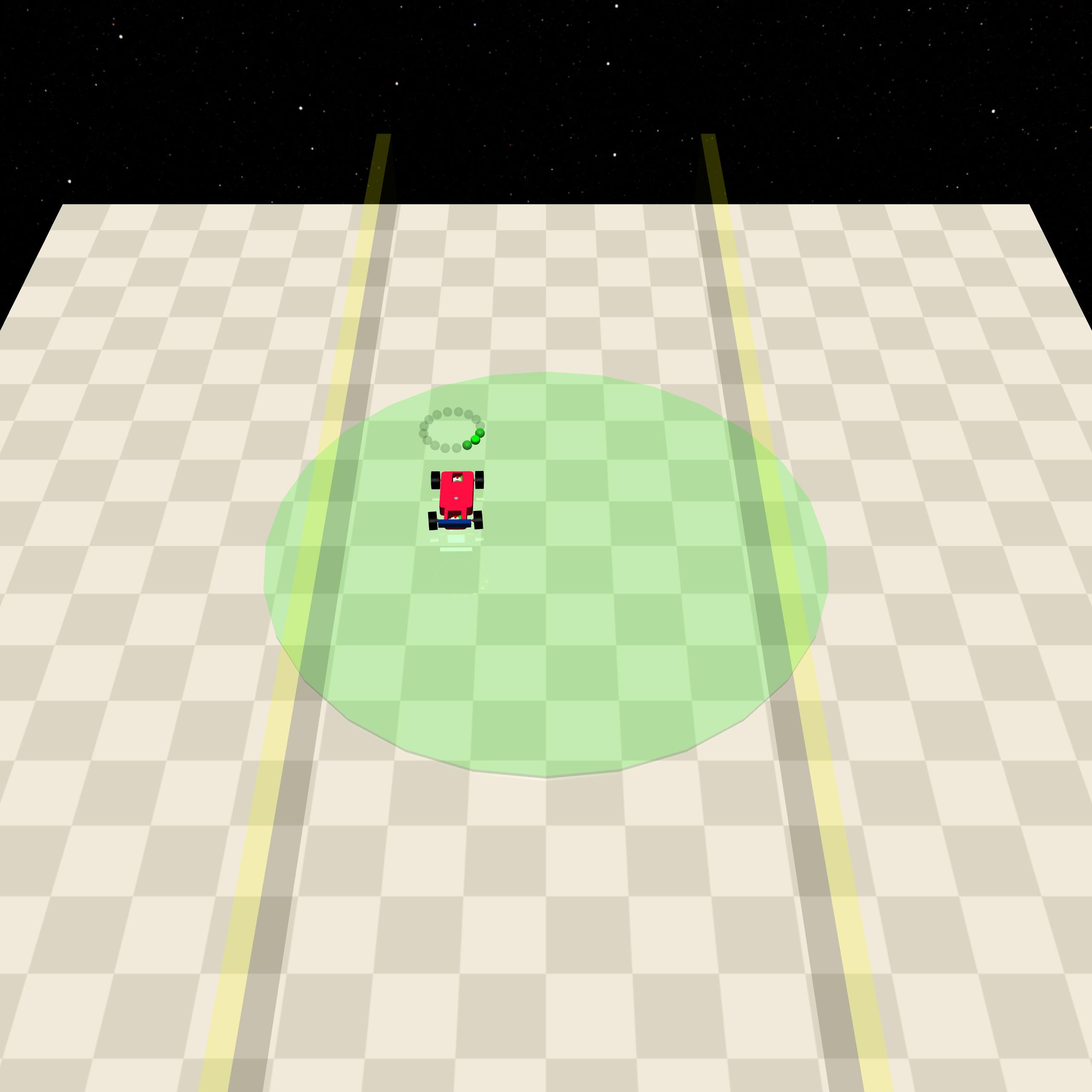}
        \caption{Circle 1}
    \end{subfigure}\hfill
    \begin{subfigure}{0.3\linewidth}
        \centering
        \includegraphics[width=\linewidth]{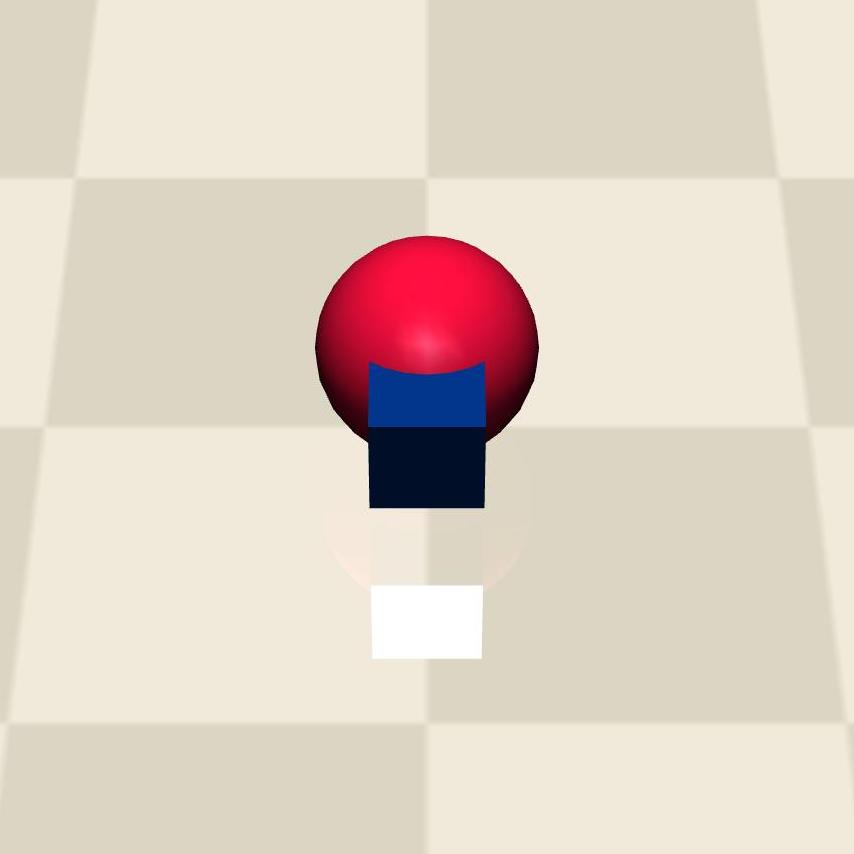}
        \caption{Point}
    \end{subfigure}\hfill
    \begin{subfigure}{0.3\linewidth}
        \centering
        \includegraphics[width=\linewidth]{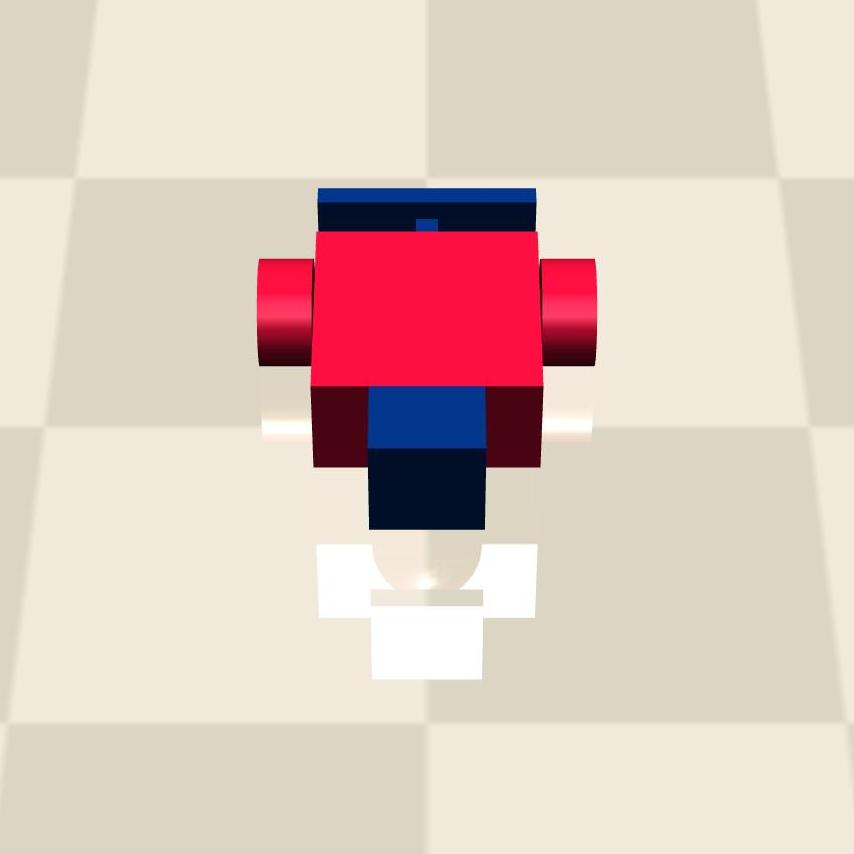}
        \caption{Car}
    \end{subfigure}\hfill
    \caption{Safety Circle Agents}
\label{fig:safe_agents}
\end{figure}

Safety Gymnasium provides several environments; we focus on the circle environment. There are 3 circle tasks with progressively difficult constraints (Fig.~\ref{fig:safe_agents}). All involve the agent moving around in a circle, but the agent incurs a cost of 1 for every step while outside a constraint boundary.
There are also several agents with different control complexities (Fig.~\ref{fig:safe_agents}).

\subsection{Results}
\begin{figure}
    \centering
    \begin{subfigure}{0.24\textwidth}
        \centering
        \includegraphics[width=\textwidth]{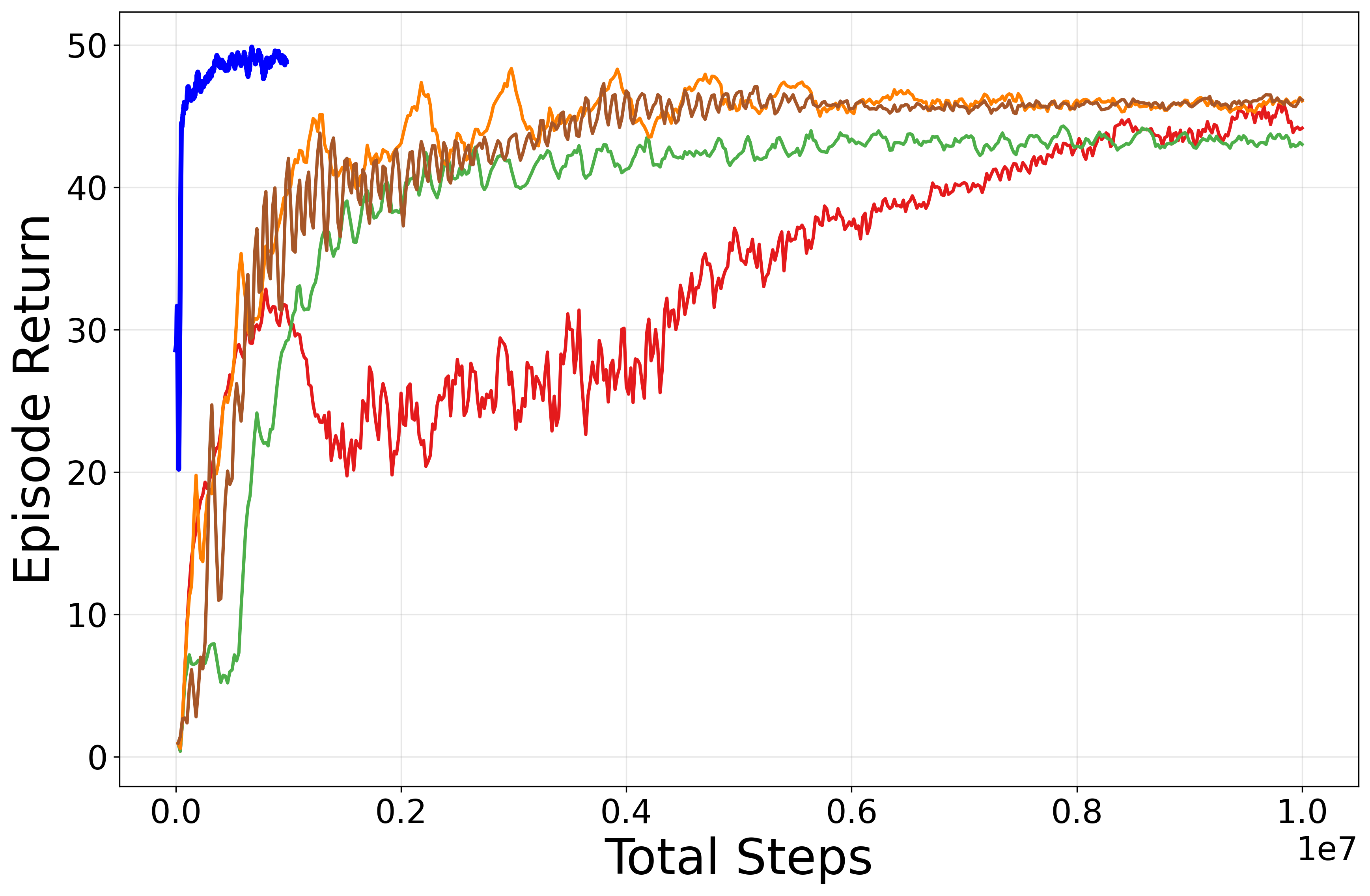}
        \caption{Point, Circle1, Reward}
    \end{subfigure}
    \begin{subfigure}{0.24\textwidth}
        \centering
        \includegraphics[width=\textwidth]{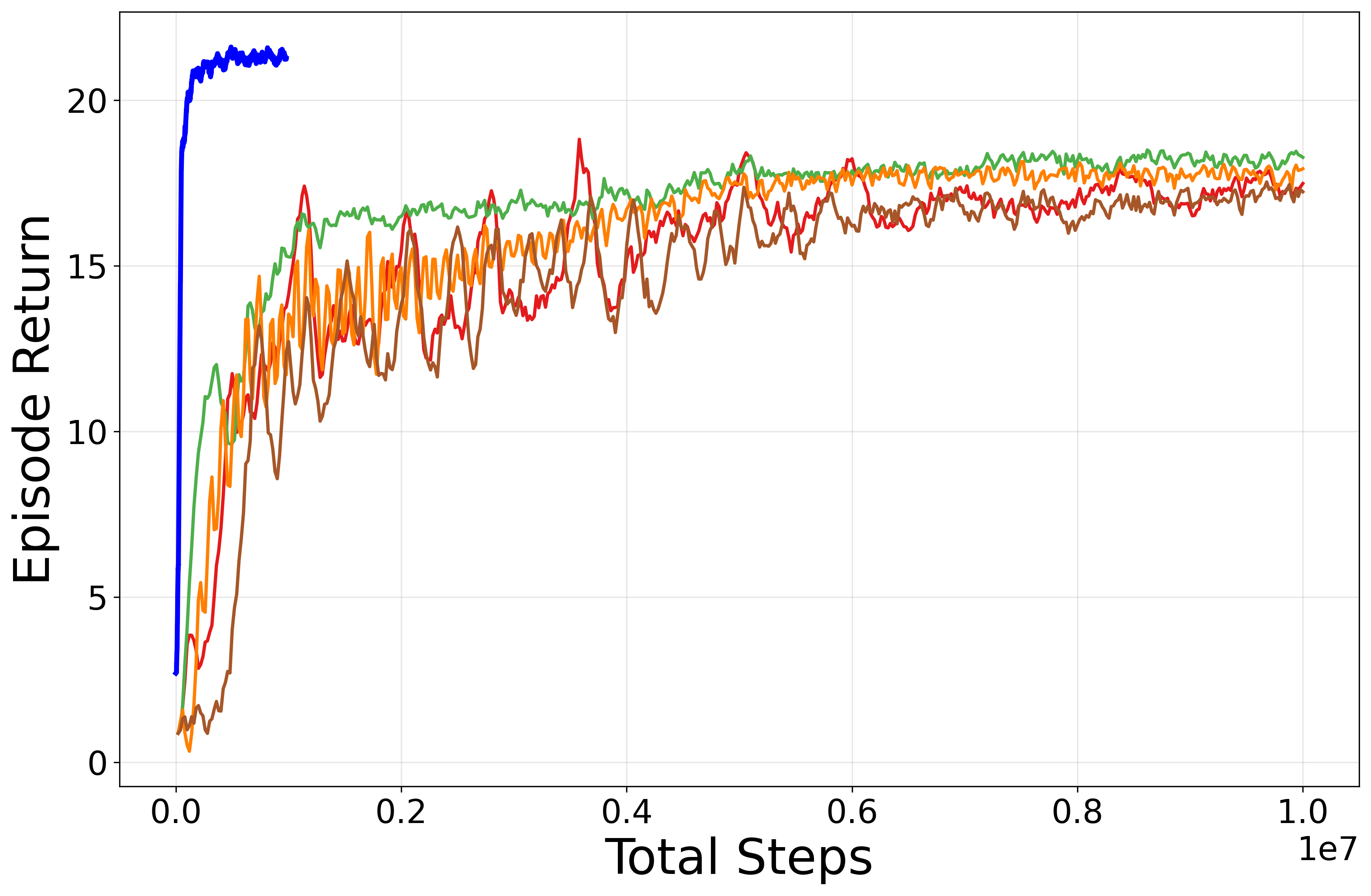}
        \caption{Car, Circle1, Reward}
    \end{subfigure}\hfill

     \begin{subfigure}{0.24\textwidth}
        \centering
        \includegraphics[width=\textwidth]{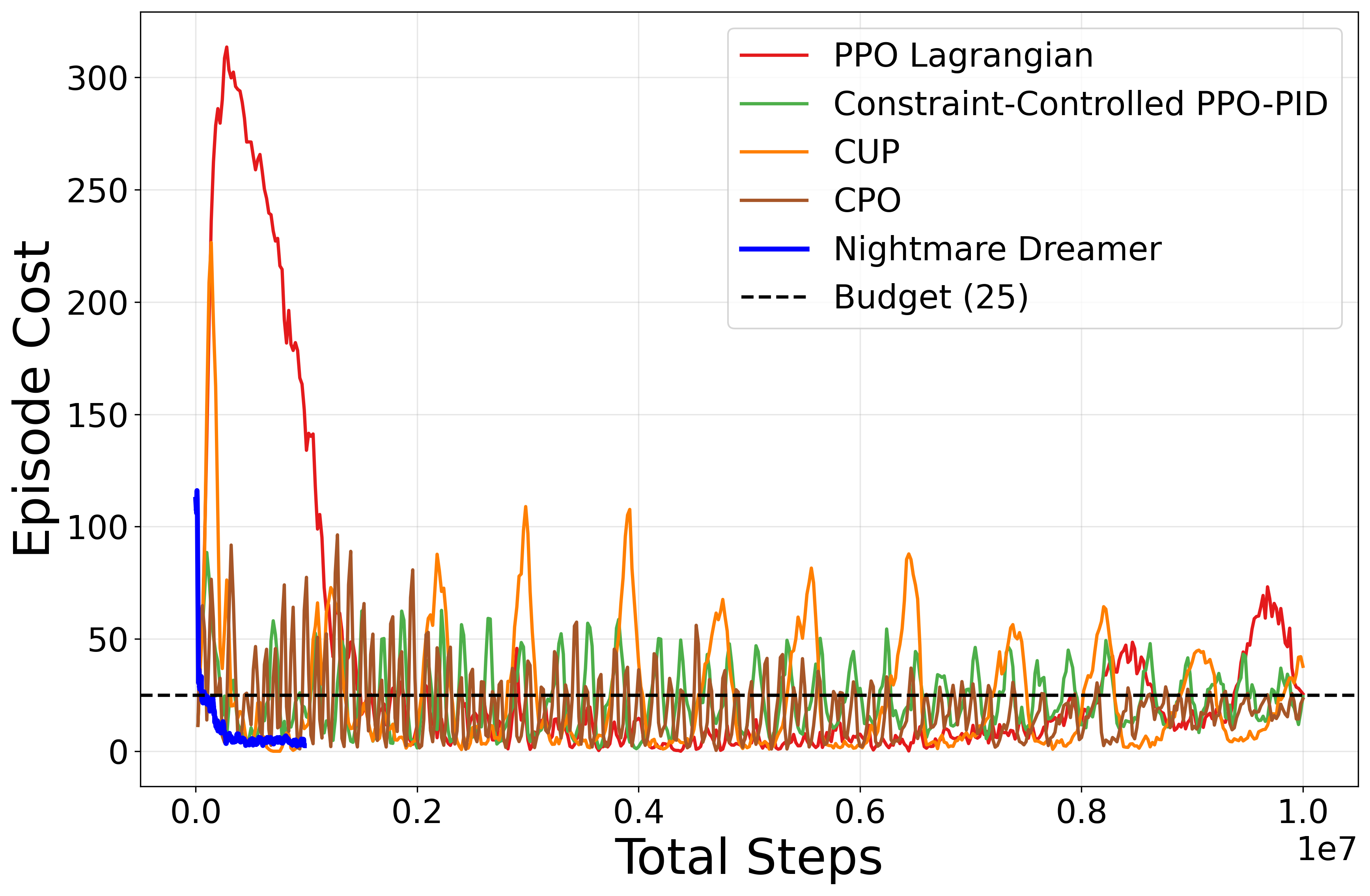}
        \caption{Point, Circle1, Cost} 
    \end{subfigure}
     \begin{subfigure}{0.24\textwidth}
        \centering
        \includegraphics[width=\textwidth]{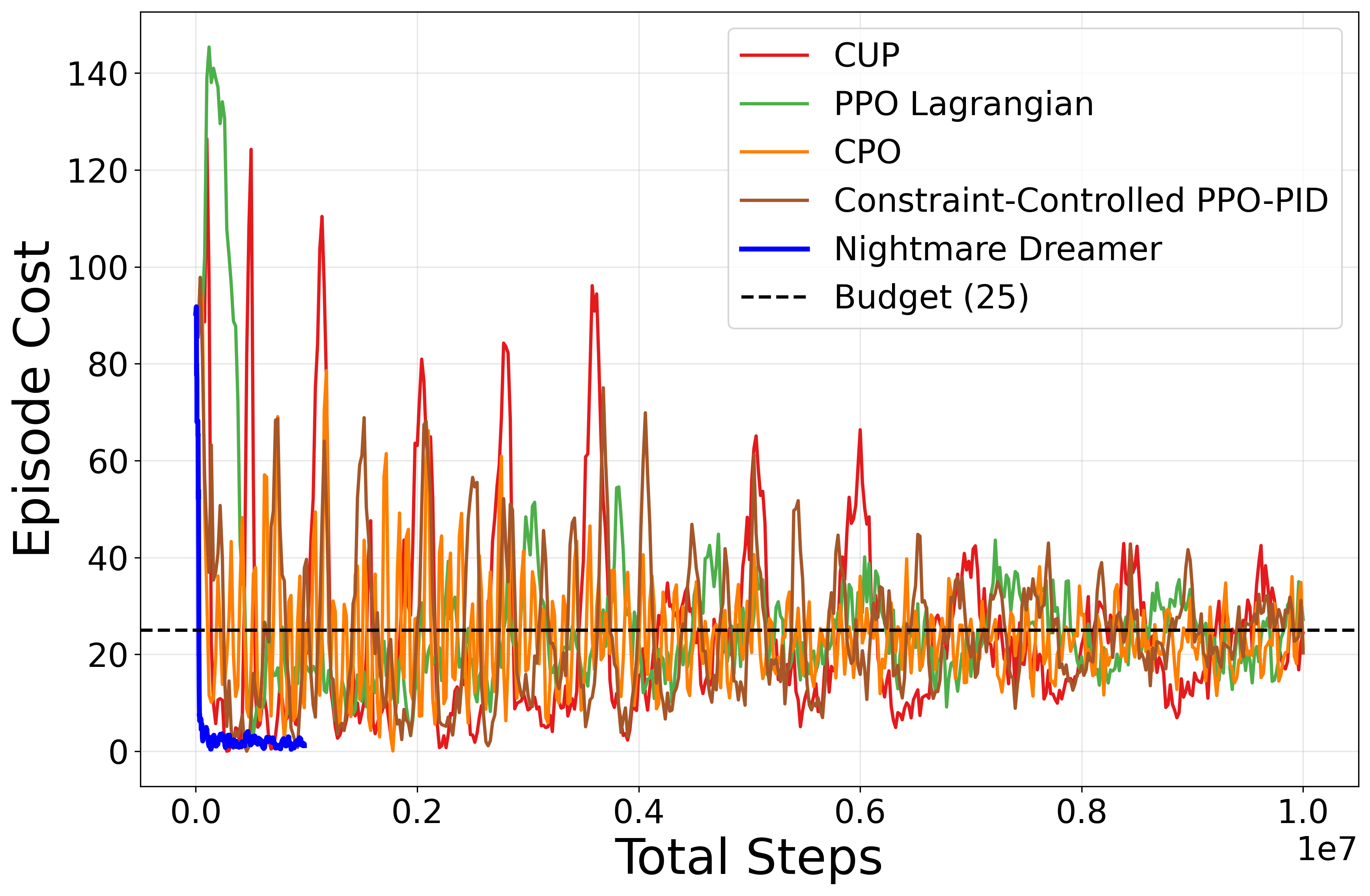}
        \caption{Car, Circle1, Cost}
    \end{subfigure}
    \caption{Safety Circle 1 Reward and Cost Performance Comparison with Benchmark algorithms}
\label{fig:results}
\end{figure}

Figure~\ref{fig:results} shows experimental results of our algorithm on SafePointCircle and SafeCarCircle compared to SOTA SafeRL algorithms from the SafePo benchmark. We observe promising results, with Nightmare converging to optimal reward and cost faster than benchmark algorithms. The dashed line indicates the safety budget that cost values must remain below. Nightmare stays below this budget while achieving near-zero cost. Due to Nightmare's computational requirements, we compare 1e6 environmental interactions against 1e7 for benchmark methods.
We compare our approach to SOTA model-free algorithms that are more computationally efficient than our approach but significantly less sample efficient, similar to the Safe-Dreamer \cite{huang2024safedreamersafereinforcementlearning} evaluation. 

\section{Conclusion} 
\label{sec:conclusion}
We introduced \textit{Nightmare Dreamer}, a model-based safe RL algorithm that achieves zero constraint violations while maximizing rewards from visual inputs. Our method trains two specialized actors—control and safe—using a learned world model, with a planning mechanism that switches between policies based on predicted future costs. Our key innovations is discriminator-based policy regularization approach. Experiments on Safety Gymnasium's Circle task demonstrate faster convergence to safe policies compared to model-free baselines. While currently validated on Circle tasks, the framework provides a foundation for extending model-based safe RL to more complex environments.
In the future, we hope to adapt \textit{Nightmare Dreamer} to other tasks and to perform tests with real-world robots and constraints and with comparison to other model-based Safe RL methods.

\balance
\bibliographystyle{plainnat}
\bibliography{references}

\end{document}